\documentclass[utf8]{FrontiersinHarvard} 

\usepackage[onehalfspacing]{setspace}
\usepackage{url,hyperref,lineno,microtype}
\usepackage{graphics}
\usepackage{csquotes}
\usepackage{siunitx}
\DeclareSIUnit{\wtp}{wt\%}
\DeclareSIUnit{\volp}{vol\%}
\DeclareSIUnit{\atp}{at\%}
\DeclareSIUnit{\um}{\micro\metre}
\DeclareSIUnit{\nm}{\nano\metre}
\DeclareSIUnit{\nms}{\micro\metre\squared}
\DeclareSIUnit{\GPa}{\giga\pascal}

\usepackage{multirow}
\usepackage{xcolor}
\usepackage{todonotes}  
\newcommand{\Stefan}[1]{%
	\todo[linecolor=red!80!black, backgroundcolor=red!60!black!5, bordercolor=red!80!black, inline]%
	{\fontsize{9}{10}\selectfont\sffamily\textbf{Stefan: }  #1}%
}

\newcommand{\Ruth}[1]{%
	\todo[linecolor=green!80!black, backgroundcolor=green!60!black!5, bordercolor=green!80!black, inline]%
	{\fontsize{9}{10}\selectfont\sffamily\textbf{Ruth: }  #1}%
}

\newcommand{\Chen}[1]{%
	\todo[linecolor=blue!80!black, backgroundcolor=blue!60!black!5, bordercolor=blue!80!black, inline]%
	{\fontsize{9}{10}\selectfont\sffamily\textbf{Chen: }  #1}%
}
\newenvironment{add-note}{\par\color{blue}}{\par}

\usepackage[noabbrev, capitalise, sort&compress]{cleveref}    
\Crefname{figure}{Figure}{Figures}
\Crefname{table}{Table}{tables}

\newcommand{\Bmu}{{\ensuremath{\boldsymbol{\mu}}}}
\newcommand{\Btheta}{{\ensuremath{\boldsymbol{\theta}}}}
\newcommand{\BSigma}{{\ensuremath{\boldsymbol{\Sigma}}}}
\newcommand{\BPhi}{{\ensuremath{\boldsymbol{\Phi}}}}

\DeclareMathAlphabet\rsfscr{U}{rsfso}{m}{n}
\newcommand{\calD}{{\ensuremath{\rsfscr{D}}}}
\newcommand{\calL}{{\ensuremath{\mathcal{L}}}}
\newcommand{\trainingDS}{{\ensuremath{\calD}^{\textrm train}}}
\newcommand{\testingDS}{{\ensuremath{\calD}^{\textrm test}}}

\usepackage{tcolorbox}
\tcbuselibrary{listings}

\newtcblisting{foo}{
  listing only,
  nobeforeafter,
  after={\xspace},
  hbox,
  tcbox raise base,
  fontupper=\ttfamily,
  colback=lightgray,
  colframe=lightgray,
  size=fbox
  }

\def\keyFont{\fontsize{8}{11}\helveticabold }
\def\firstAuthorLast{Zhang {et~al.}} 
\def\Authors{Chen Zhang\,$^{1}$, Cl\'{e}mence Bos\,$^{2}$, Stefan Sandfeld\,$^{1,3}$, and Ruth Schwaiger\,$^{4, 5, \ast}$}
\def\Address{
     $^{1}$Institute for Advanced Simulation -- Materials Data Science and Informatics (IAS-9), 
    Forschungszentrum J\"ulich GmbH, J\"ulich, Germany \\
    $^{2}$Institute for Applied Materials, Karlsruhe Institute of Technology (KIT), Eggenstein-Leopoldshafen, Germany \\ 
    $^{3}$Chair of Materials Data Science and Informatics, 
    Faculty of Georesources and Materials Engineering, RWTH Aachen University, Aachen, Germany\\
    $^{4}$Institute of Energy and Climate Research -- Structure and Function of Materials (IEK-2),     Forschungszentrum J\"ulich GmbH, J\"ulich, Germany \\
    $^{5}$Chair of Energy Engineering Materials, 
    Faculty of Georesources and Materials Engineering, RWTH Aachen University, Aachen, Germany 
}

\def\corrAuthor{Ruth Schwaiger}
\def\corrEmail{r.schwaiger@fz-juelich.de}

\begin{document}
\onecolumn
\firstpage{1}

\title {Unsupervised Learning of Nanoindentation Data to Infer Microstructural Details of Complex Materials} 

\author[\firstAuthorLast ]{\Authors} 
\address{} 
\correspondance{} 

\extraAuth{}

\maketitle

\begin{abstract}

In this study, Cu-Cr composites were studied by nanoindentation. Arrays of indents were placed over large areas of the samples resulting in datasets consisting of several hundred measurements of Young's modulus and hardness at varying indentation depths. The unsupervised learning technique, Gaussian mixture model, was employed to analyze the data, which helped to determine the number of ``mechanical phases'' and the respective mechanical properties. Additionally, a cross-validation approach was introduced to infer whether the data quantity was adequate and to suggest the amount of data required for reliable predictions -- one of the often encountered but difficult to resolve issues in machine learning of materials science problems.

\tiny\keyFont{\section{Keywords:} 
    unsupervised learning, cross-validation, Gaussian mixture model,
    CuCr composite, mechanical properties, nanoindentation} 
\end{abstract}

\section{Introduction}
\label{sec1}

Nanoindentation is a powerful technique for characterizing the mechanical properties of materials at small length scales\,\citep{oliver2004:JMR19, liu2018:JPSE167}.
For analyzing nanoindentation data, data analysis and machine learning techniques have gained importance in recent years. A CNN-based classifier\,\citep{kossman2021:M14} was developed to identify whether pop-in events are present in the load-displacement curves from nanoindentation tests. The classifier achieved an accuracy of approximately 93\%, which helped understand the process that created pop-ins. In another study\,\citep{vignesh2019:MD181}, the phase level features were extracted from spatial hardness and elastic modulus maps using a deconvolution method based on the k-means clustering algorithm. Furthermore, a Gaussian mixture model (GMM)\,\citep{sorelli2008:CCR38} was employed to develop the fiber-reinforced ultra-high performance concrete. To characterize the nano-mechanical properties of the phases governing the microstructure, the experimental probability distribution functions (PDFs) of indentation modulus and indentation hardness were deconvolved; this unsupervised learning approach has also proven beneficial for mapping the mechanical properties of complex surfaces in two dimensions by deconvolving the experimental cumulative distribution functions of indentation modulus and hardness for naval brass, cast iron, Ti64-10TiC alloy, and M3 high speed steel\,\citep{randall2009:JMR24}.

The GMM technique can be implemented in a wide range of materials science fields. It has been applied to the automated analysis and visualization of continuum fields in atomistic simulations, where individual grain's distributions of total strain, elastic strain, and rotation are extracted as key features\,\citep{prakash2022:AEM24}. It has also been used to analyze the so-called \emph{grain orientation spread} from electron back-scattered diffraction, which measures intragranular lattice distortion. Three zones were separated from the GMM results, representing the recrystallized zone, mixture zone and relict zone, eliminating the ambiguity in the selection of the cut-off threshold when identifying the recrystallized grains\,\citep{yeo2023:JSG}. It has also been applied to high resolution high-angle annular dark-ﬁeld scanning transmission electron microscopy data on counting the number of atoms of monotype crystalline nanostructures, assuming that the total scattered intensity is proportional to the number of atoms per atom column\,\citep{de2013:U134}. In addition, it has been utilized in X-ray diffraction investigations, enabling automatic extraction of charge density wave order parameters and detection of intraunit cell ordering and its fluctuations from a series of high-volume X-ray diffraction measurements taken at multiple temperatures\,\citep{venderley2022:PNAS119}. Using nanoindentation data to investigate the distribution of heterogeneous materials has a great potential, provided the GMM technique is applied properly and a sufficient amount of data is available.

In this work, the properties of the phases of two different Cu-Cr composites and the volume fractions of the constituent phases were explored using a clustering technique based on the Gaussian mixture model, and the model results were validated. Initial tests in this study were conducted at a depth of \SI{1}{\um} on single-phase pure Cu and pure Cr. Our methodology was then applied at the same depth to Cu-Cr composites with a nominal Cr content of 25 and 60 wt\%.  Additional datasets were included in the analysis to investigate the effect of data size on model performance.

\section{Materials and Methodology}
\label{sec2}
\subsection{Experimental details}\label{subsec2-1} 
\begin{figure}[htbp]
    \centering
    \includegraphics[width=0.8\textwidth]{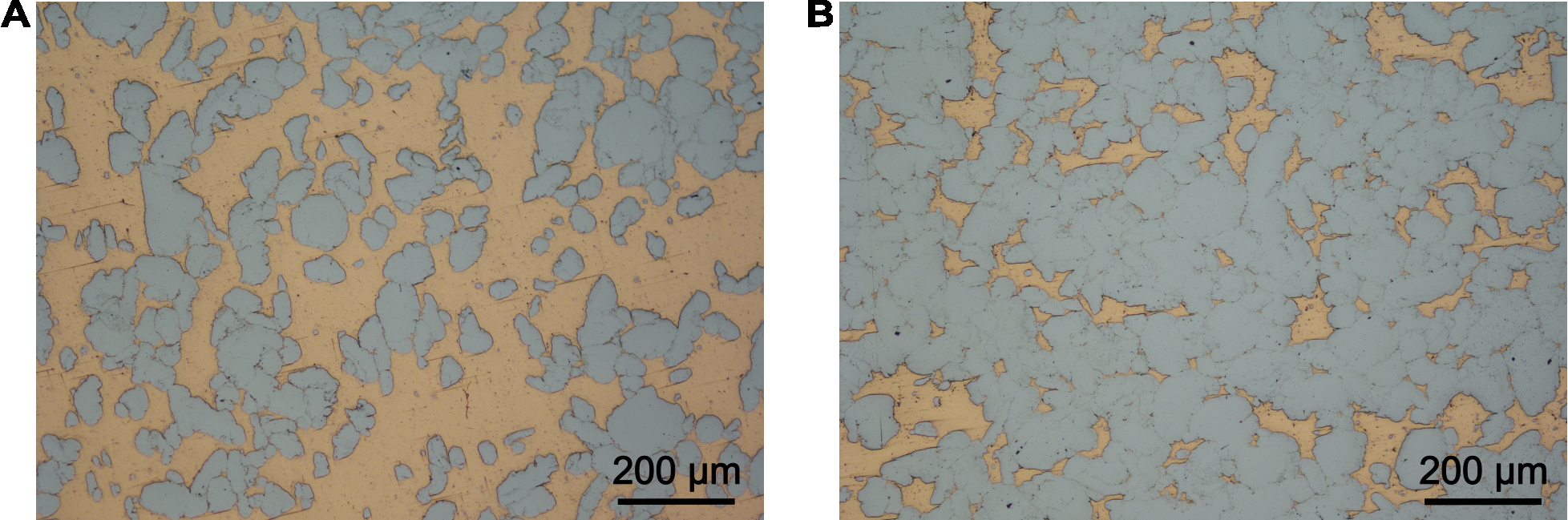}
    \caption{Optical micrographs of Cu-Cr composites produced by field-assisted sintering technique with (a) \SI{25}{\wtp} Cr, (referred to as CuCr25),  and (b) \SI{60}{\wtp} Cr (referred to as CuCr60) were investigated applying the grid indentation technique.  
    }
    \label{fig:CuCr_samples_1}
\end{figure}

In this study, we have evaluated different Cu-Cr composites containing \SI{25}{\wtp} Cr and \SI{60}{\wtp} Cr corresponding to \SI{29.95}{\atp} and \SI{64.40}{\atp} Cr, respectively, as well as Cu and Cr as reference samples. All materials were produced via field-assisted sintering technique (FAST) as described in detail in \citep{vonKlinski2015:TUDthesis}. Briefly, Cu powder with \SI{99.9}{\atp} purity and technically pure Cr powder (\SI{99.5}{\atp}) were used and compacted at a temperature of \SI{950}{\celsius} and a pressure of 40 MPa, except for the Cr sample that was compacted at a temperature of \SI{1450}{\celsius}. The composite samples (~\cref{fig:CuCr_samples_1}) will be referred to as CuCr25 and CuCr60 according to their nominal compositions. For indentation testing, the sample surfaces were prepared using standard grinding and polishing techniques using SiC paper and diamond suspensions with decreasing grain size down to \SI{0.1}{\um}. 

%

A nanoindenter G200 XP (Agilent/Keysight Technologies, Inc., CA, USA) equipped with a diamond Berkovich tip was used to investigate the mechanical properties of the composite samples. The samples were indented to different depths using the so-called Express Test option. Arrays of indents covering areas up to \SI{500}{\um} x \SI{500}{\um} were made in different locations on the sample surface. The indentation depths ranged from \SI{300}{\nm} to \SI{1000}{\nm} and the distance between individual indents was between \SI{20}{\um} and \SI{23}{\um} for all depths. Hardness $H$ and Young's modulus $E$ were determined assuming \SI{1141}{\GPa} and 0.07 for Young's modulus and Poisson's ratio, respectively, of the diamond tip.

\begin{figure}[htbp]
    \centering
    \includegraphics[width=0.5\textwidth]{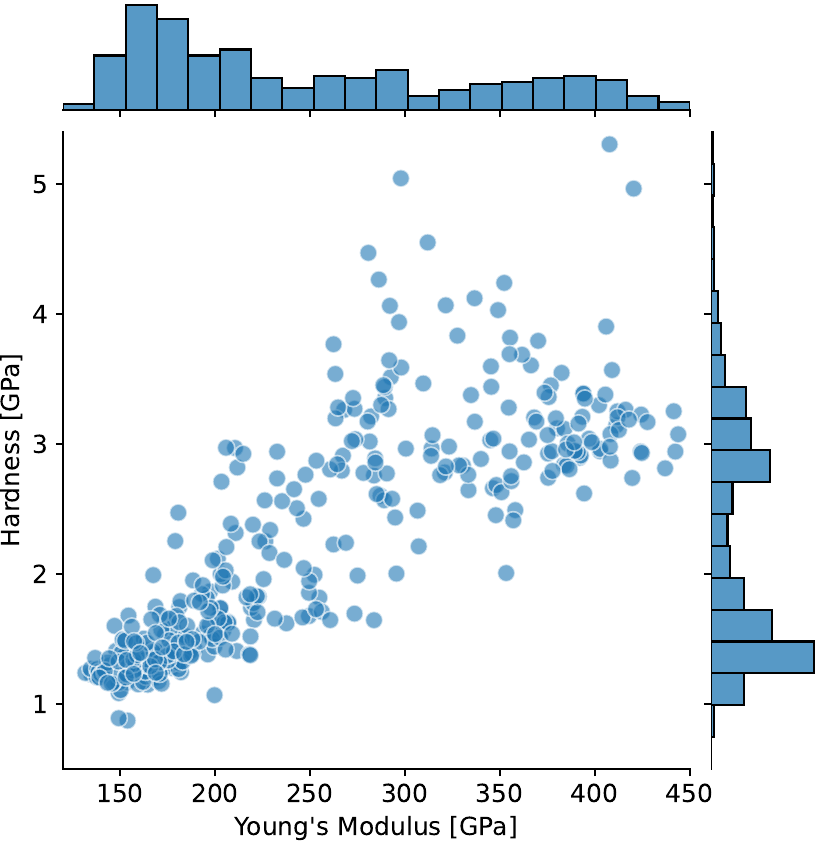}
    \caption{Scatter plot with marginal histograms of all obtained values for Young's modulus and hardness measured for CuCr60. 
    }
    \label{fig:histogram}
\end{figure}

 \cref{fig:histogram} shows an example distribution of hardness and Young's modulus of CuCr60 for an indentation depth of \SI{1}{\um}. The marginal hardness distribution is a bimodal distribution, while a trimodal distribution can be seen for the marginal histogram of Young's modulus. 



%
%

\subsection{Clustering based on the Gaussian mixture model}
\label{subsec2-2}

The Cu-Cr composites consisted of two distinct phases with average Cr particle diameters of approximately \SI{30}{\um} on the indented surface; thus, upon indenting the surface we assume that either one or the other element dominates or the properties of a mixture of both phases is measured. However, the \enquote{mixture of elements} might as well be more than \enquote{just the sum of its parts} since additional effects, e.g., related to the presence of interfaces, might occur during indentation. Furthermore, we assume that similar local microstructural or chemical properties lead to similar measurement data and that the mechanical properties exhibit gradual changes over the surface. 

Thus, the data scientific task is to analyze a number of data records consisting of the given feature variable $E$ (Young's modulus) and the feature $H$ (hardness), 
given for four different materials and at different depths, i.e., \SI{300}{\nm}, \SI{400}{\nm} and \SI{1000}{\nm}. The whole dataset $\mathcal{D}$ consisting of $N$ data records can be written as the set of pairs  
\begin{align}
    \calD = \left\{(E_i, H_i)\right\}_{i=1\ldots N}\;.
\end{align}
The goal is to perform a clustering analysis, during which data points with similar elastic properties are grouped together, i.e., for each pair $(E_i, H_i)$ in the feature space, we determine the categorical variable $y_i$ that gives the number of the respective cluster. 
Additionally, the total number of different clusters needs to be
determined. Since annotated training data does not exist, this has do be done in an unsupervised manner.

For the clustering, we use the Gaussian mixture model (GMM), which is a robust and well-established probabilistic clustering model in the statistics literature (see, e.g., \citep{bishop2006:S4, reynolds2009:EB741, reynolds1995:ITSAP3}). Each different \enquote{phase} is assumed to correspond to an individual Gaussian distribution of Young's Modulus and hardness. These materials \enquote{phases} are commonly referred to as \emph{components} in the context of machine learning. We assume that the distribution of experimental data was generated by a combination of Gaussian processes, which are represented by the probability density functions (PDFs) $\mathcal{N}$ for each component $j$. The resulting superposition is then given by
\begin{align} \label{eq:PDF}
p\left(\calD; \BPhi\right)=\sum_{j=1}^{k} \alpha_{j} \mathcal{N}\left(\calD \mid \Btheta_{j}\right)\;,
\end{align}
where 
$k$ is the number of components of the model, $\alpha_{j} > 0$ are the weights of each 
component $j$, the $\Btheta_j$ are the vectors of parameters for the Gaussian
and $\BPhi=\{\alpha_{1}, ..., \alpha_{k}, \Btheta_{1}, ..., \Btheta_{k}\}$ is a short
notation for the whole set of parameters governing the Gaussian mixture model. 
For a multivariate Gaussian, the component $j$ of the superimposed function is given by the parameters  $\Btheta_j = \{\Bmu_j, \BSigma_j\}$, where $\Bmu_j$ and $\BSigma_j$ are the mean value and the covariance matrix, respectively. $\Bmu_j$ describes the location of the component $j$ in the feature space, while  the covariance matrix $\BSigma_j$ characterizes the $j$-th component data distributed around $\Bmu_j$. The objective of the training process is to estimate the values for the model parameters of the Gaussian distribution that best align with the training data $\trainingDS \subset \calD$. The model parameters $\BPhi_{k}$ are iteratively determined while assuming the number of clusters $k$ to be predefined. Typically, the superposition of Gaussians is computed using the maximum likelihood method based on a set of candidate models that differ in the number of clusters generated using the expectation maximization algorithm. For further details of the algorithm used, please refer to the appendix section\,\ref{appendix: MLE}.

The Bayesian Information Criterion ($\operatorname{BIC}$) is employed as a selection criterion to identify the optimal model. It is widely recognized and applied in model selection\,\citep{gideon1978:TAS6}, providing a measure for assessing the accuracy of the unsupervised GMM:
\begin{equation}\label{eq: BIC}
\operatorname{BIC} = -2 \ln \calL + d \ln N
\end{equation}
where $d$ denotes the number of parameters of the model, and $\calL$ is the maximum likelihood achieved by the model used. The first term represents the maximized likelihood of a model, 
and the second term introduces a penalty for the number of parameters to mitigate the risk of overfitting. The model with the lowest $\operatorname{BIC}$ value indicates the highest likelihood and better predictive capability for the observed data. In this study, the GMM has been implemented using the open-source Python package scikit-learn\,\citep{scikit-learn2011:JMLR}.

\section{Results and Discussion}
\label{sec3}

In the following, we will start with the data cleaning process and the analysis of pure metals using a 1D Gaussian mixture model. Then, the CuCr25 and CuCr60 composites are investigated using both 1D and 2D GMM for an indentation depth of \SI{1}{\um}, with a detailed discussion of the impact of sample size on model robustness. Finally, we will evaluate the effect of indentation depth on the 1D GMM results for the depth range of \SI{200}{\nm} to \SI{600}{\nm}.

\subsection{Preparation of the datasets}\label{subsec3-1}
\begin{figure}[ht!]
    \centering
    \includegraphics[width=0.9\textwidth]{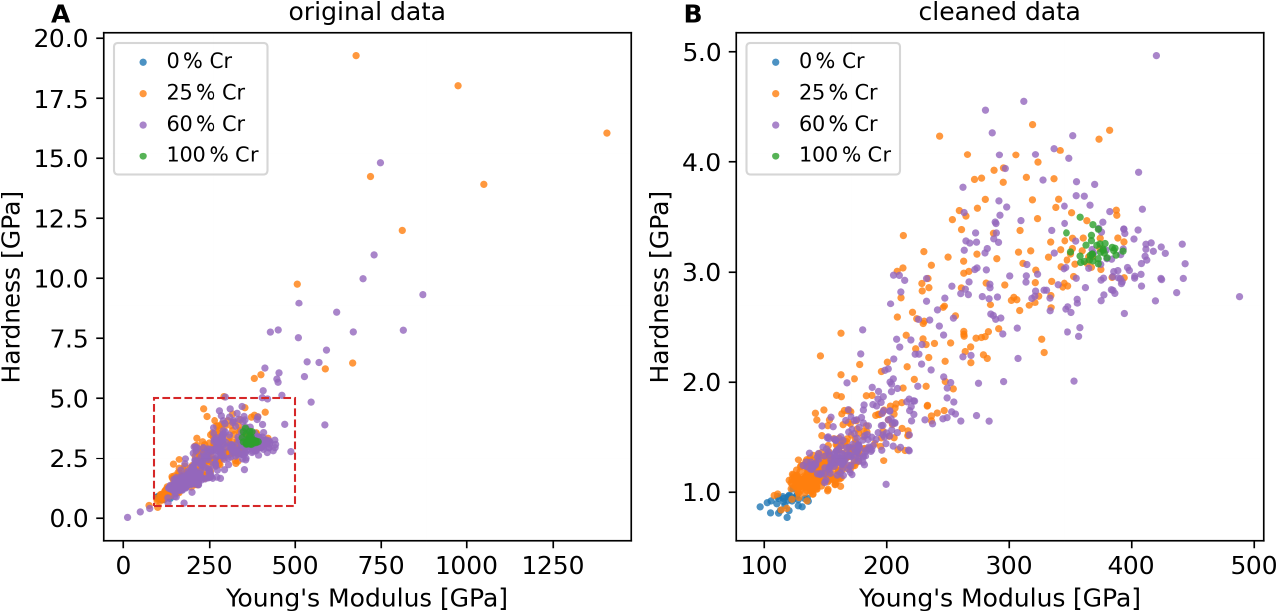}
    \caption{%
        The experimentally obtained distributions of Young's modulus and hardness
        for \SI{0}{\wtp}, \SI{25}{\wtp}, \SI{60}{\wtp} and \SI{100}{\wtp} Cr content.
        (\textbf{A}) The experimental data in its original form, (\textbf{B}) The 
        cleaned and preprocessed data set used in our analysis. The rectangle in \textbf{A} represents the region illustrated in \textbf{B}.
    }
    \label{fig:four_alloys_EG_distribution}
\end{figure}

\Cref{fig:four_alloys_EG_distribution}$\textbf{A}$ illustrates the original data sets comprising the measured $E$ and $H$ values of the four materials tested. 
Due to variations in the height of compacted particles on the polished surface, the actual recorded maximum indentation depths ranged from \SI{600}{\nm} to \SI{2000}{\nm}, deviating from the nominal \SI{1000}{\nm}. We focused our analysis on depths ranging from \SI{800}{\nm} to \SI{1200}{\nm}. Subsequently, the data underwent cleaning and filtering processes to remove measurement errors, particularly outliers with unrealistically high values as well as other invalid data. For CuCr25, data within the ranges of $\SI{100}{} \leq E \leq \SI{400}{\GPa}$ and $\SI{0.8}{} \leq H \leq \SI{4.5}{\GPa}$ were retained, while for CuCr60 the data range was $\SI{100}{} \leq E \leq \SI{500}{\GPa}$ and $\SI{1.0}{} \leq H \leq \SI{5.0}{\GPa}$. The resulting cleaned data, comprising approximately 98\% of the original data, is presented in \Cref{fig:four_alloys_EG_distribution}$\textbf{B}$. Notably, the distributions of the two pure metals exhibit lower variances compared to the Cu-Cr composites. Further preprocessing of the data, such as standardization, was found to have no significant impact on the training results and were therefore not included in this study.
%
%
%


\subsection{Mechanical properties of the pure Cu and pure Cr specimens}
\label{subsec3-2}
\begin{figure}[ht!]
    \centering
    \includegraphics[width=\textwidth]{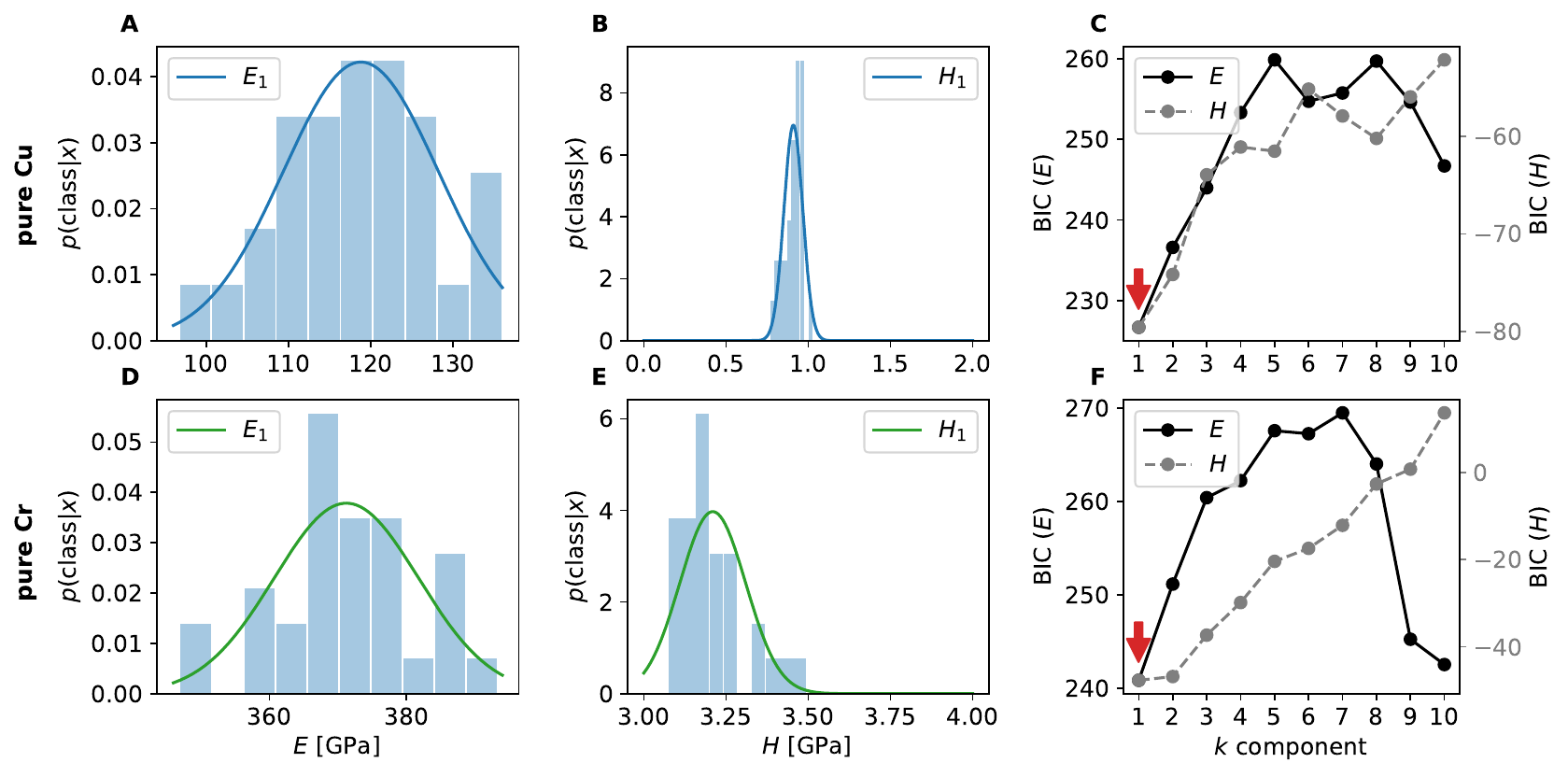}
    \caption{
        Probability density functions (left and middle column) and plot of 
        BIC (right column) of pure Cu and Cr. The BIC values are shown for both, 
        Young's modulus (left ``$y$'' axis) and hardness (right ``$y$'' axis).
        The top row shows the data for Cu, the bottom row for Cr.
    }
    \label{fig:pure_Cu_Cr}
\end{figure}

The mechanical properties of pure Cu and pure Cr were analyzed for the indentation depth of \SI{1}{\um}. The PDF plots of $E$ and $H$ are depicted in \cref{fig:pure_Cu_Cr}$\textbf{A}$-$\textbf{E}$ showing the mean values of $E=\SI{118.80}{\GPa}$ and $H=\SI{0.91}{\GPa}$ for pure Cu, and $E=\SI{371.24}{\GPa}$ and $H=\SI{3.21}{\GPa}$ for pure Cr. The bin size was selected to ensure an approximately equal number of bins covering the range of all PDFs. To utilize the GMM, it is important to demonstrate that the distributions are roughly normally distributed, despite the GMM's inherent robustness. Thus, the Shapiro-Wilk test was employed to assess the normality of the distributions\,\citep{oztuna2006:TJMS36}, resulting in $p$-values of 0.95 for $E$ and 0.14 for $H$ in pure Cu, and 0.95 for $E$ and 0.14 for $H$ in pure Cr. The test accepts the normality hypothesis, when the $p$-value exceeds 0.05, confirming that the data are sufficiently normally distributed. The optimal number of components was 1 for both the Cu and Cr specimens (as shown in \cref{fig:pure_Cu_Cr}$\textbf{C}$ and \cref{fig:pure_Cu_Cr}$\textbf{F}$), as determined by the BIC analysis. Regarding determination of the component numbers, the result of 1D GMM for $E$ with high $p$-values is therefore more reliable than the fit of $H$. Nonetheless, because $H$ is strongly correlated with local microstructure details, such as grain boundaries and dislocation pile-ups\,\citep{oliver1992:JMR7, oliver2004:JMR19}, the fitting result of $H$ can be used to infer local microstructural characteristics. This can also be seen from the small variation at the beginning of the BIC plot for hardness in \cref{fig:pure_Cu_Cr}$\textbf{F}$, where the BIC values for $k=2$ differ only marginally from the one at $k=1$ that we had identified as the optimal one.

\subsection{Mechanical properties of the CuCr25 and CuCr60 specimens}
\label{subsec3-3}

\begin{figure}[ht!]
    \centering
    \includegraphics[width=\textwidth]{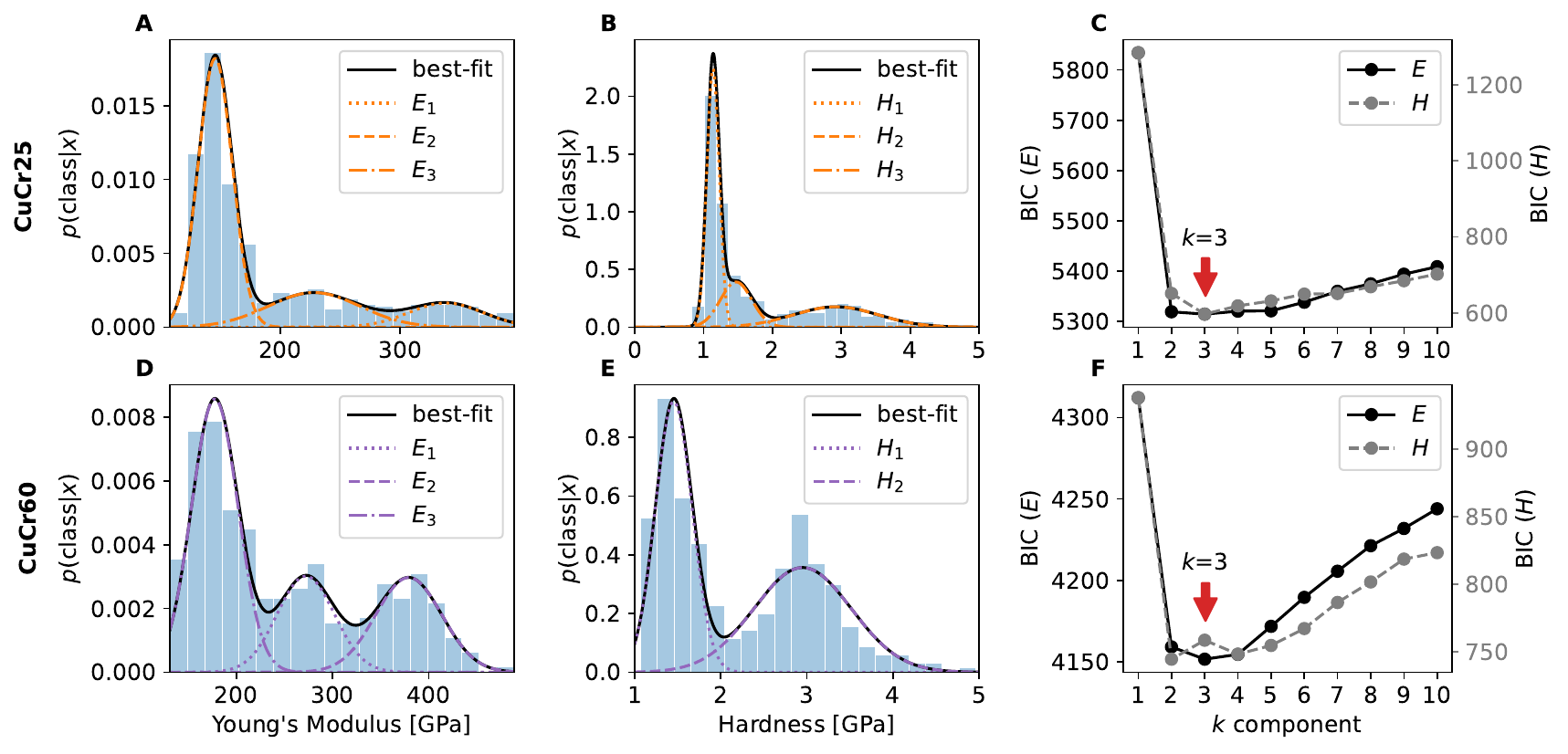}
    \caption{1D GMM results of CuCr25 and CuCr60 at \SI{1}{\um} indentation depth. CuCr25: \textbf{(A)} histogram of $E_i$ and the best fit (solid line), \textbf{(B)} histogram of $H_i$ and the best fit (solid line), \textbf{(C)} BIC of $H$ and $E$; CuCr60: \textbf{(D)} histogram of $E_i$ and the best fit model, \textbf{(E)} histogram of $H_i$ and the best fit model, \textbf{(F)} BIC of $H$ and $E$.}\label{fig:CuCr_1D-GMM}
\end{figure}

The histograms of $E$ and $H$, 
together with their best-fit models are shown in \cref{fig:CuCr_1D-GMM}$\textbf{A}$ and \cref{fig:CuCr_1D-GMM}$\textbf{D}$ (left panel) and \cref{fig:CuCr_1D-GMM}$\textbf{B}$ and \cref{fig:CuCr_1D-GMM}$\textbf{E}$ (middle panel). The right panel (\cref{fig:CuCr_1D-GMM}$\textbf{C}$ and \cref{fig:CuCr_1D-GMM}$\textbf{F}$) shows the BIC as a function of the number of components. 

The BIC values for both the modulus and the hardness cover a range of around $500$, and even $\approx 200$ for hardness, excluding $k=1$. Though not obvious, this is an important difference. 
Taking CuCr60 for instance, there are $N\approx 300$ valid data for  $E$ determined at a depth of \SI{1}{\um}. Assuming an optimal number of clusters of $k=3$, the number of parameters to be measured in a 1-dimensional analysis are (i) three mean values ($\mu_1, \mu_2, \mu_3$), (ii) three variance values ($\sigma_1^2, \sigma_2^2, \sigma_3^2$), and (iii) two coefficients that determine the relative weights of the three Gaussians ($\alpha_1, \alpha_2$). Given that the sum of all weights should be 1, $\sum_{j=1}^k \alpha_j=1$, and $\alpha_3 = 1 - \left(\alpha_1+ \alpha_2\right)$, the number of parameters to be determined is $d=8$. 

Are the differences in BIC values large or small? To answer this question, we need to understand how much variation results from a small change in the dataset. The calculated values of the logarithm of $\mathcal{L}$ (cf. \cref{eq: BIC}) of this dataset are between $-10$ and $-4$. According to \cref{eq: BIC}, the first term varies between $-2\times -4=8$ and $-2\times -10=20$ and the second term is $8\times \ln{300}\approx45$. The sum of the two terms lies between 53 and 65. This indicates, assuming a minor variation in 1 out of 300 data points, that the BIC variation should fall between 53 and 65. In other words, the BIC very effectively captures changes in the data. If the BIC difference between the two models for this dataset is greater than $\approx 50$, it already suggests that the corresponding number of components $k$ is indeed the more likely one. In this example, the difference between a BIC value for $k$ and $k+1$ is slightly smaller. However, taking a look at larger values of $k$, there is a very clear trend that points to the minimum. 
In conclusion, our 1D GMM analysis of Young's modulus reveals that both alloys have three mechanical phases at \SI{1}{\um} depth. \cref{tab:1D and 2D GMM at 1 micrometer depth} summarizes the results of the Gaussian mixture models evaluated for the four different materials.

%

As stated above, our GMM analysis of CuCr25 shows the presence of three mechanical phases as well as differences in the properties of nominally identical phases. In CuCr25, for example, the Cu-rich phase accounts for \SI{64.5}{\volp} (defined by $E_1$ as the lowest value, closest to the value of pure Cu), while $E_2$ and $E_3$ account for a total of \SI{35.5}{\volp}. These results indeed coincide well with the \SI{35.5}{\volp} Cr estimated by optical microscopy \citep{bos2019:KITthesis}. The lowest modulus value for the Cu phase was found in the compacted Cu sample (i.e. $E_1$), followed by $E_1$ of CuCr60 and CuCr25. 
$E_3$ is Young's modulus of the Cr-rich phase (with $E_3$ being the highest, closest to the modulus value of pure Cr).  The highest modulus of the Cr phase (i.e. $E_1$) was found for the compacted Cr sample, followed by the $E_3$ values of the CuCr60 and the CuCr25 samples.

These differences in the mechanical properties fitted by GMM are likely related to the diffusion of one phase into the other, the presence of foreign particles or pores, or the influence of the surrounding material. Assuming that a small amount of Cr, i.e. \SI{0.4}{\atp} - \SI{3}{\atp}, can be dissolved in the Cu matrix \citep{Jacob2000:ZMetall, Chakrabarti1984:BullAllPhaDiag}, the Cr solid solution likely contributes to the difference of the modulus of the Cu phase in the composite samples. In addition, the presence of Cr nanoparticles of 100 - 200 nm in size was reported \citep{vonKlinski2015:TUDthesis}, which as well results in a higher Young's modulus value of the Cu phase in CuCr25 and CuCr60. Finally, considering the relative densities of 99.4 \% and 98.3 \% for CuCr25 and CuCr60, respectively, compared to 97.7 \% for the pure Cu sample, a Young's modulus value with an estimated reduction up to $9\%$ \citep{Lebedev1995:MSEA203} can be expected. The reduction of the modulus of the Cr phase in CuCr25 (i.e. $E_3$) can be explained by the surrounding softer Cu phase. While in general also the hardness fitting of CuCr25 supports the presence of three mechanical phases, the percentage results are different. This is not surprising, though, since hardness is a local property that is strongly related to the microstructure including grain and phase boundaries as well as dislocations.

	\begin{table}[ht!]\footnotesize\centering
		\caption{1D mechanical property fitting based on the optimal BIC results at \SI{1}{\um} depth}\label{tab:1D and 2D GMM at 1 micrometer depth}%
            {%
			\begin{tabular}{clcccc}
				\hline
				\multirow{2}{*}{Cr {[}wt\%{]}} & \multirow{2}{*}{$\mathrm{F}_{3}$/$\mathrm{F}_{2}$/$\mathrm{F}_{1}$/$\mathrm{F}_{0}$} & \multicolumn{4}{c}{1D}   \\ \cline{3-6}  
				                               &          
                                               & \begin{tabular}[c]{@{}c@{}}Mechanical\\ property\end{tabular} 
                                               & \begin{tabular}[c]{@{}c@{}}Average\\ {[}GPa{]}\end{tabular} 
                                               & \begin{tabular}[c]{@{}c@{}}Standard\\ deviation\end{tabular} 
                                               & \begin{tabular}[c]{@{}c@{}}Percentage\\ {[}\%{]}\end{tabular} \\ \hline
				\multirow{2}{*}{0 (pure Cu)}   & \multirow{2}{*}{30/30/33/35} & $E_1$   & 118.80   & 9.45   & 100   \\
				                               &                              & $H_1$   & 0.91     & 0.06   & 100   \\ \hline
				                               &                              &         &          &        &       \\
				\multirow{7}{*}{25 (CuCr25)}   &                              & $E_1$   & 145.55   & 14.12  & 64.5  \\
				                               &                              & $E_2$   & 226.50   & 38.42  & 22.6  \\
				                               &                              & $E_3$   & 337.02   & 31.69  & 12.9  \\
				                               & 513/520/555/675              &         &          &        &       \\
				                               &                              & $H_1$   & 1.14     & 0.09   & 50.1  \\
				                               &                              & $H_2$   & 1.47     & 0.23   & 22.6  \\
				                               &                              & $H_3$   & 2.92     & 0.62   & 27.3  \\ \hline
				                               &                              &         &          &        &       \\
				\multirow{6}{*}{60 (CuCr60)}   &                              & $E_1$   & 177.35   & 24.17  & 52.2  \\
				                               &                              & $E_2$   & 272.17   & 29.75  & 22.4  \\
				                               &                              & $E_3$   & 379.32   & 34.34  & 25.4  \\
				                               & 366/376/427/675              &         &          &        &       \\
				                               &                              & $H_1$   & 1.45     & 0.22   & 50.0  \\
				                               &                              & $H_2$   & 2.96     & 0.56   & 50.0  \\ \hline
				                               &                              &         &          &        &       \\
				\multirow{2}{*}{100 (pure Cr)} & \multirow{2}{*}{31/35/35/35} & $E_1$   & 371.24   & 10.54  & 100   \\
				                               &                              & $H_1$   & 3.21     & 0.10   & 100   \\ \hline
			\end{tabular}%
		}
	\end{table}

	\begin{table}[ht!]\footnotesize\centering
		\caption{2D mechanical property fitting based on the optimal BIC results at \SI{1}{\um} depth}\label{tab:2D GMM at 1 micrometer depth}%
              {%
			\begin{tabular}{clccc}
				\hline
                \multirow{2}{*}{Cr {[}wt\%{]}} & \multirow{2}{*}{$\mathrm{F}_{3}$/$\mathrm{F}_{2}$/$\mathrm{F}_{1}$/$\mathrm{F}_{0}$} & \multicolumn{3}{c}{2D}  \\\cline{3-5} 
				                               &                                                                                      & \begin{tabular}[c]{@{}c@{}}Average $E$\\ {[}GPa{]}\end{tabular} 
                                                                                                                                      & \begin{tabular}[c]{@{}c@{}}Average $H$\\ {[}GPa{]}\end{tabular} 
                                                                                                                                      & \begin{tabular}[c]{@{}c@{}}Percentage\\ {[}\%{]}\end{tabular} \\ \hline
                                               &                  &  144.91 & 1.17 & 61.1 \\
                25 (CuCr25)                    & 513/520/555/675  &  220.92 & 2.25 & 27.1 \\
                                               &                  &  340.52 & 3.18 & 11.8 \\\hline%
                                               &                  & 172.32  & 1.42  & 43.4 \\
                60 (CuCr60)                    & 366/376/427/675  & 262.63  & 2.67  & 34.6 \\
                                               &                  & 383.35  & 3.02  & 22.0 \\
                \hline%
			\end{tabular}%
		}
	\end{table}

\begin{figure}[ht!]
   \centering
   \includegraphics[width=\textwidth]{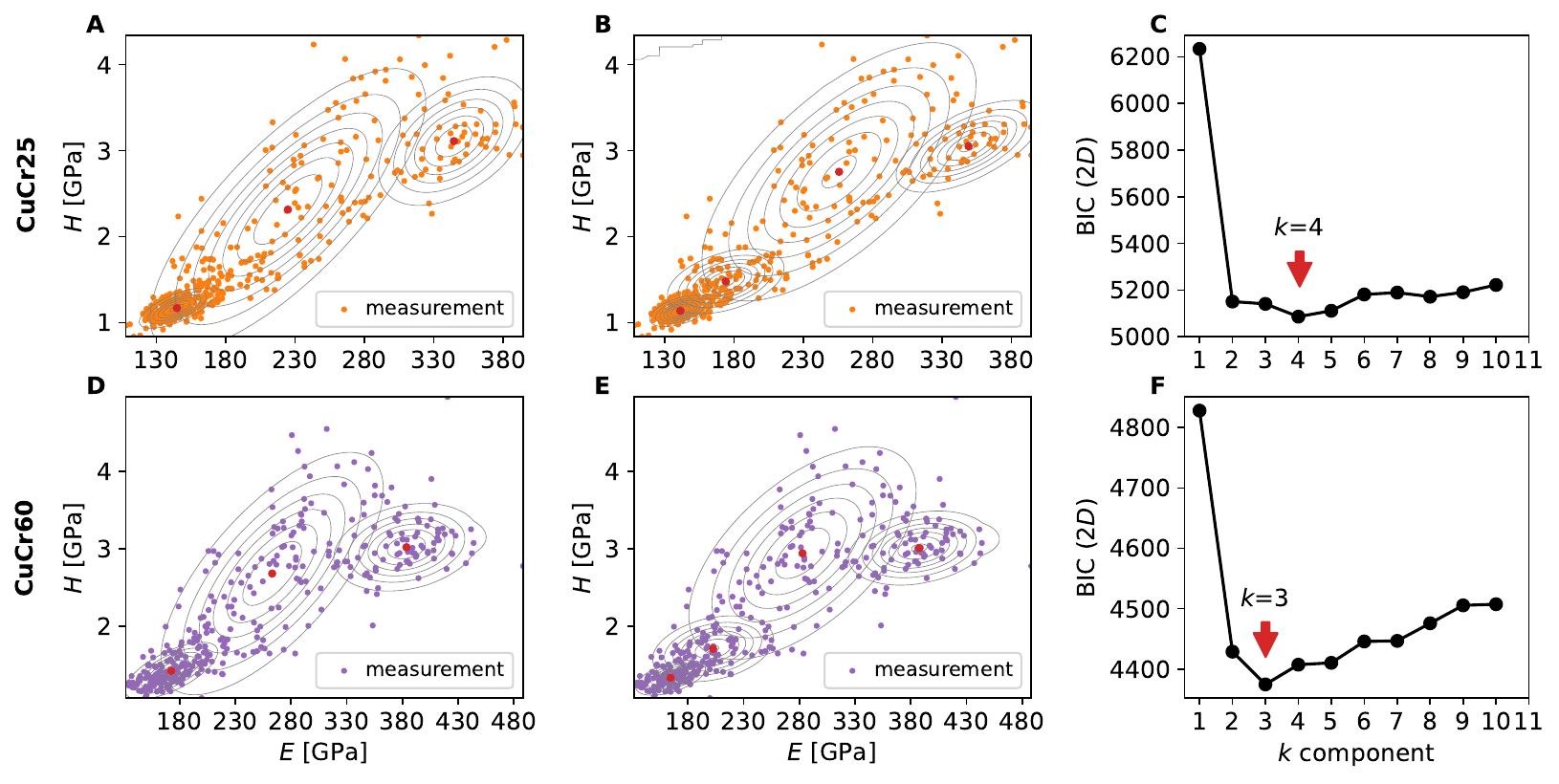}
   \caption{2D Gaussian mixture model clustering of CuCr25 and CuCr60. CuCr25: {(A)} Three components. {(B)} Four components and \textbf{(C)} 
   2D BIC; CuCr60: {(D)} Three components {(E)} Four components and {(F)} 2D BIC. The ellipses in A, B, D, and E are isolines of the 
   Gaussian distributions and the red points represent the average values of the different components.
   }
   \label{fig:CuCr_2D-GMM}
\end{figure}

After discussing the 1D GMM analysis, we now take a look at the outcomes of the 2D GMM with independent feature variables $E$ and $H$ shown in \cref{fig:CuCr_2D-GMM}. The distribution of three and four components or mechanical phases of CuCr25 and CuCr60, respectively, are shown in \cref{fig:CuCr_2D-GMM}$\textbf{A}$-$\textbf{B}$ and \cref{fig:CuCr_2D-GMM}$\textbf{D}$-$\textbf{E}$. The red points in the graphs represent the average value of each component, while the concentric ellipses with different orientations (covariance) represent the different components. The model selection criteria BIC are shown as a function of the number of components in the right panel (\cref{fig:CuCr_2D-GMM}$\textbf{C}$ and \cref{fig:CuCr_2D-GMM}$\textbf{F}$). 

Based on \cref{fig:CuCr_2D-GMM}$\textbf{C}$, one could say that four mechanical phases are the ideal match for the CuCr25 composite. Given the anomaly in the upper left corner of \cref{fig:CuCr_2D-GMM}$\textbf{B}$, though, the best assumption remains at three, which will be further discussed in \cref{subsec3-4}. As shown in \cref{tab:1D and 2D GMM at 1 micrometer depth}, in the 2D GMM, which combines both $E$ and $H$, the estimated amount of Cr in CuCr25 was \SI{38.9}{\volp}, which is close to the actual experimental findings \citep{bos2019:KITthesis}.

The 2D GMM analysis (in \cref{fig:CuCr_2D-GMM}$\textbf{F}$) reveals that also CuCr60 contains three mechanical phases at \SI{1}{\um} depth. The fitted result for the volume of Cr in 1D is \SI{47.8}{\volp} based on the modulus values, which is less than the nominal value (i.e., \SI{65}{\volp} Cr). By contrast, the 2D GMM result indicates a Cr volume fraction of \SI{56.6}{\volp} (\cref{tab:1D and 2D GMM at 1 micrometer depth}). The difference between the 1D GMM ($E$) and 2D GMM results is related to the $H$, while the difference between the fitting results and the experimental data is due to the amount of data and variation of microstructures over the samples. Note that the number of datapoints for CuCr60 is 40\% less than that for CuCr25; insufficient data can be the source of inaccuracies, which we are addressing in the following. 

\begin{figure}[ht!]
    \centering
    \includegraphics[width=\textwidth]{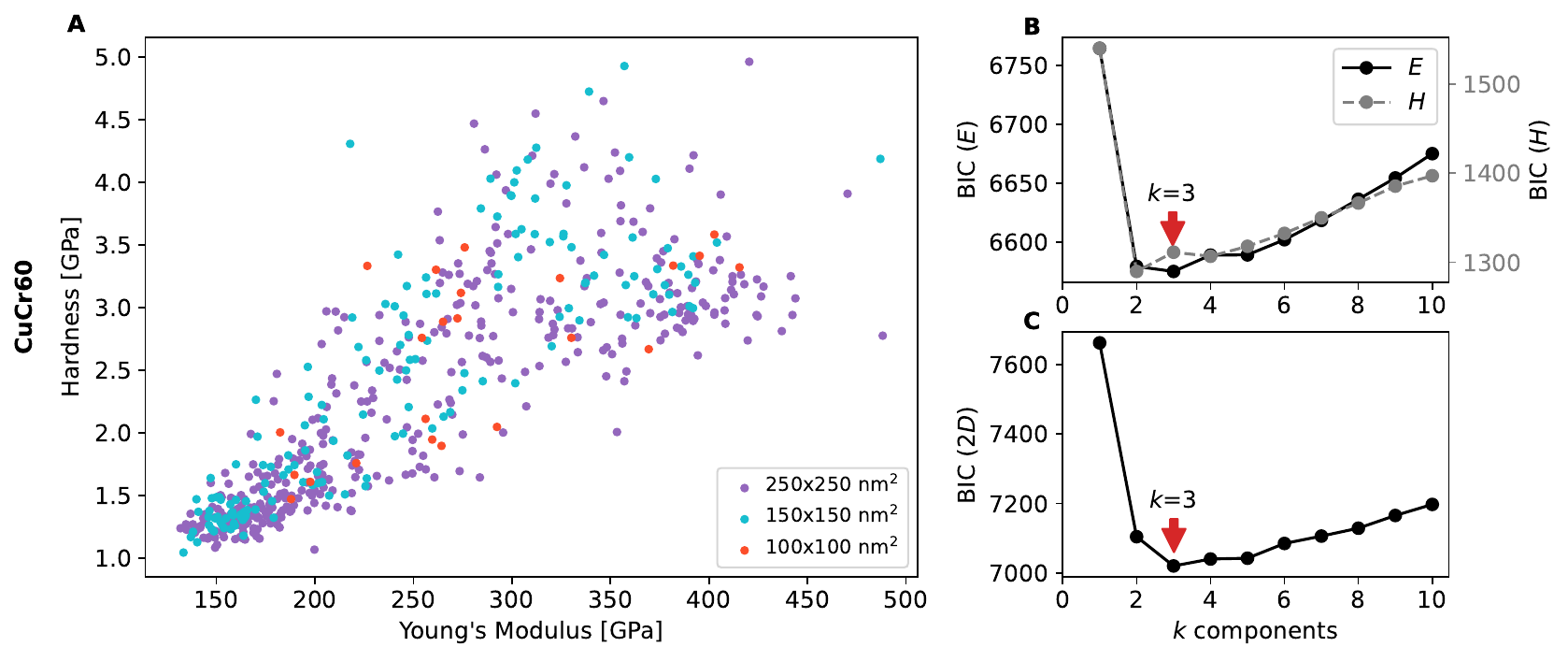}
    \caption{%
    (A) Distribution of the CuCr60 data over the depth range of \SI{500}{\nm} - \SI{2000}{\nm} (combined datasets for indentation arrays of \SI{100}{\um}\,x\,\SI{100}{\um}, \SI{300}{\um}\,x\,\SI{300}{\um}, and  \SI{500}{\um} \,x\, \SI{500}{\um} size) \textbf{(B)} 1D BIC for $E$ and $H$, and \textbf{(C)} 2D BIC for $E$ and $H$.}\label{fig:CuCr60_1D_2D-GMM}
\end{figure}

In the above analysis, the size of the dataset was 1,844 and collected over an area of \SI{500}x\,\SI{500}{\nms}. To increase the size of the dataset, we merged it with two more nanoindentation areas (\SI{100}x\,\SI{100}{\nms} and \SI{300}x\,\SI{300}{\nms}) and analyzed the data using the same procedures described above. The results are summarized in \cref{tab:CuCr60_three_indent_areas1D} and \cref{tab:CuCr60_three_indent_areas2D}. Merging the three datasets now includes indentation depths ranging from \SI{500} {\nm} to almost \SI{2000} {\nm} with 97.8\% of the data lying between \SI{800}{}-\SI{1200}{\nm} indentation depth. As shown in \cref{fig:CuCr60_1D_2D-GMM}$\textbf{A}$, the distribution ranges of the data coincide well indicating that the microstructure of the material was comparable over the different areas indented. 

As shown in \cref{fig:CuCr60_1D_2D-GMM}$\textbf{B}$, the most likely number of phases determined by analyzing Young's Modulus is three, while two phases are most probable when analyzing the hardness. The 2D fit, though, also indicates $k=3$, as can be seen in \cref{fig:CuCr60_1D_2D-GMM}$\textbf{C}$.  The individual 2D fits for the different areas of \SI{300}x\,\SI{300}{\nms} and the \SI{500}x\,\SI{500}{\nms}, each reflect three mechanical phases with \SI{67.1}{\volp} Cr and \SI{59.5}{\volp} Cr, respectively. Not unexpectedly, the merged data then yielded a Cr fraction of \SI{61.5}{\volp}. The somewhat different results reflect the variation of the microstructures over the sample surface and underscores the importance of identifying characteristic microstructures or increasing the size of the dataset.

    \begin{table}[ht!]
        \centering\footnotesize
        \caption{1D mechanical property fitting of CuCr60 based on the optimal BIC results at \SI{1}{\um} depth}\label{tab:CuCr60_three_indent_areas1D}%
        {%
        \begin{tabular}{ccccc}
        \hline
        \multirow{2}{*}{\begin{tabular}[c]{@{}c@{}}Indentation \\ area\\  {[}\SI{}{\nms}{]}\end{tabular}} & \multicolumn{4}{c}{1D}                                                                                                                                                                                                                               \\ \cline{2-5} 
                                                                                                  & \begin{tabular}[c]{@{}c@{}}Mechanical \\ property\end{tabular} & \begin{tabular}[c]{@{}c@{}}Average\\ {[}GPa{]}\end{tabular} & \begin{tabular}[c]{@{}c@{}}Standard \\ deviation\end{tabular} & \begin{tabular}[c]{@{}c@{}}Percentage\\ {[}\%{]}\end{tabular} \\ \hline
        \multirow{4}{*}{100x100}                                                                  & $E_9$                                                          & -                                                           & -                                                             & -                                                             \\
                                                                                                  &                                                                &                                                             &                                                               &                                                               \\
                                                                                                  & $H_1$                                                          & 1.83                                                        & 0.21                                                          & 39.1                                                          \\
                                                                                                  & $H_2$                                                          & 3.15                                                        & 0.29                                                          & 60.9                                                          \\ \hline
                                                                                                  &                                                                &                                                             &                                                               &                                                               \\
        \multirow{5}{*}{150x150}                                                                  & $E_1$                                                          & 157.79                                                      & 12.02                                                         & 29.2                                                          \\
                                                                                                  & $E_2$                                                          & 280.30                                                      & 70.47                                                         & 70.8                                                          \\
                                                                                                  &                                                                &                                                             &                                                               &                                                               \\
                                                                                                  & $H_1$                                                          & 1.42                                                        & 0.18                                                          & 37.3                                                          \\
                                                                                                  & $H_2$                                                          & 3.02                                                        & 0.74                                                          & 62.7                                                          \\ \hline
                                                                                                  &                                                                &                                                             &                                                               &                                                               \\
        \multirow{6}{*}{250x250}                                                                  & $E_1$                                                          & 177.86                                                      & 24.92                                                         & 49.5                                                          \\
                                                                                                  & $E_2$                                                          & 275.26                                                      & 31.88                                                         & 24.9                                                          \\
                                                                                                  & $E_3$                                                          & 378.94                                                      & 35.93                                                         & 24.6                                                          \\
                                                                                                  &                                                                &                                                             &                                                               &                                                               \\
                                                                                                  & $H_1$                                                          & 1.44                                                        & 0.22                                                          & 45.5                                                          \\
                                                                                                  & $H_2$                                                          & 2.95                                                        & 0.62                                                          & 54.5                                                          \\ \hline
                                                                                                  &                                                                &                                                             &                                                               &                                                               \\
        \multirow{6}{*}{merged data}                                                              & $E_1$                                                          & 175.12                                                      & 24.10                                                         & 46.0                                                          \\
                                                                                                  & $E_2$                                                          & 269.55                                                      & 34.10                                                         & 28.5                                                          \\
                                                                                                  & $E_3$                                                          & 373.37                                                      & 37.46                                                         & 25.5                                                          \\
                                                                                                  &                                                                &                                                             &                                                               &                                                               \\
                                                                                                  & $H_1$                                                          & 1.45                                                        & 0.22                                                          & 42.8                                                          \\
                                                                                                  & $H_2$                                                          & 2.98                                                        & 0.65                                                          & 57.2                                                          \\ \hline
        \end{tabular}%
        }
        \end{table}

        \begin{table}[ht!]
            \centering\footnotesize
            \caption{2D mechanical property fitting of CuCr60 based on the optimal BIC results at \SI{1}{\um} depth}\label{tab:CuCr60_three_indent_areas2D}%
            {%
            \begin{tabular}{cccc}
            \hline
            \multirow{2}{*}{\begin{tabular}[c]{@{}c@{}}Indentation \\ area\\  {[}\SI{}{\nms}{]}\end{tabular}} &  \multicolumn{3}{c}{2D}    \\ \cline{2-4} 
                                                                                                       & \begin{tabular}[c]{@{}c@{}}Average $E$\\ {[}GPa{]}\end{tabular} & \begin{tabular}[c]{@{}c@{}}Average $H$\\ {[}GPa{]}\end{tabular} & \begin{tabular}[c]{@{}c@{}}Percentage\\ {[}\%{]}\end{tabular} \\ \hline
            \multirow{2}{*}{100x100}                                                                   & 252.19                                                        & 2.46                                                          & 78.4                                                          \\
                                                                                                       & 393.06                                                        & 3.27                                                          & 21.6                                                          \\\hline
                                                                                                       & \multicolumn{3}{c}{}                                                                                                                                                                          \\
            \multirow{3}{*}{150x150}                                                                   & 159.47                                                        & 1.39                                                          & 32.8                                                          \\
                                                                                                       & 229.10                                                        & 2.29                                                          & 31.9                                                          \\
                                                                                                       & 337.91                                                        & 3.51                                                          & 35.2                                                          \\\hline
                                                                                                       & \multicolumn{3}{c}{}                                                                                                                                                                          \\
            \multirow{3}{*}{250x250}                                                                   & 172.25                                                        & 1.41                                                          & 40.5                                                          \\
                                                                                                       & 267.99                                                        & 2.74                                                          & 38.1                                                          \\
                                                                                                       & 383.12                                                        & 3.01                                                          & 21.4                                                          \\\hline
                                                                                                       & \multicolumn{3}{c}{}                                                                                                                                                                          \\
            \multirow{3}{*}{merged data}                                                               & 170.49                                                        & 1.42                                                          & 38.5                                                          \\
                                                                                                       & 264.73                                                        & 2.78                                                          & 40.4                                                          \\
                                                                                                       & 379.28                                                        & 3.70                                                          & 21.1                                                          \\ \hline
            \end{tabular}%
            }
            \end{table}

\subsection{Cross-validation of GMM}
\label{subsec3-4}

{Data size has been a common concern in clustering algorithms for machine learning. In contrast to the model selection criteria used in the GMM with BIC, the following discussion is intended to verify the robustness and validity of the clustering results. Clustering can be used to evaluate the effect of data size on the results and to identify the amount of experimental data required to achieve the same level of performance.} We applied the following procedure:

\begin{enumerate}

\item Given the whole dataset $\calD$, the categorical variable $y_i$ is generated by a clustering algorithm $A_k$. It constructs such a solution $\textbf{Y}:=A_k(\calD)$, where $\textbf{Y} = \{y_1, ..., y_i\}$, and $y_i=\{ 1, . . . , k\}$ is assigned to the cluster. In our case, labels $y_i$ were generated using the KMeans and GMM clustering algorithms, with the prediction of the optimal GMM algorithm serving as the ground truth.

\item $k$-fold cross-validation is conducted on the dataset $\calD$:  First, divide the data set into $k$ equal parts (the "folds"), then choose $(k-1)$ folds for training and the remaining fold for testing. Then, validate the model using the testing data set $\testingDS$ after training the model with the training data set $\trainingDS$. Perform $k$ rounds of cross-validation using multiple training datasets.

\item The adjusted Rand score is used to ensure the robustness and validity of the clustering results. A score of 1 for the adjusted Rand score indicates complete agreement between the two clusters\,\citep{chacon2023:ADAC17}.

\end{enumerate}

\begin{figure}[ht!]
\centering
\includegraphics[width=0.95\textwidth]{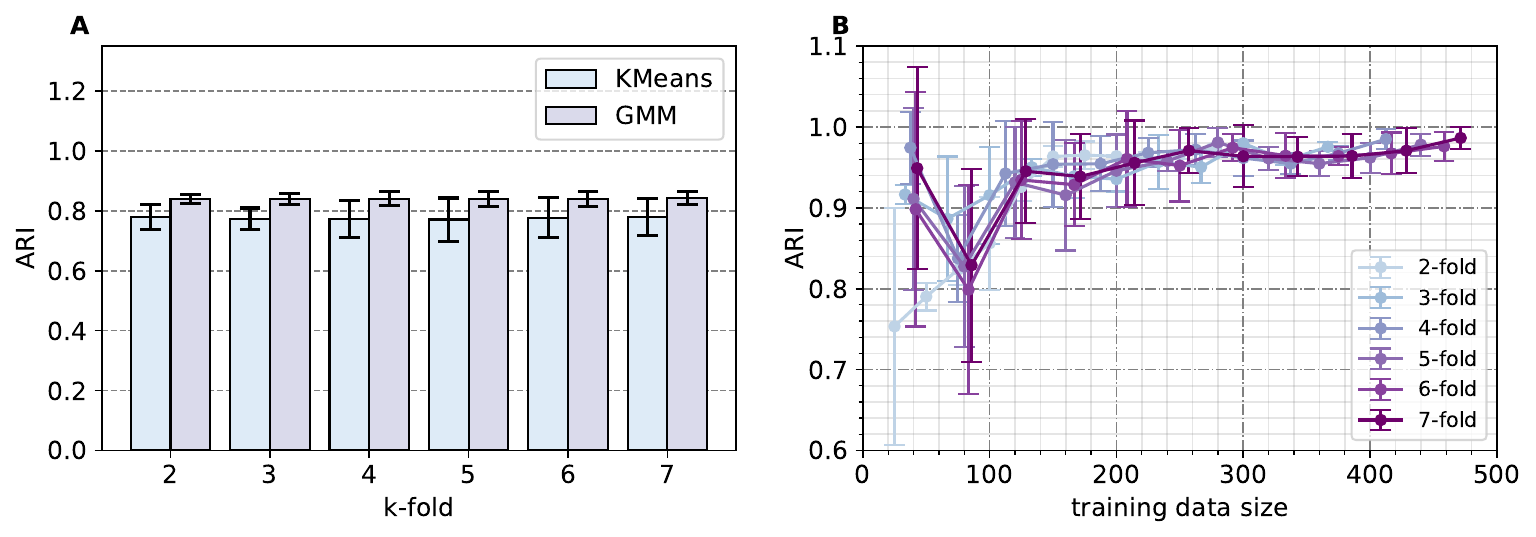}
\caption{Cross-validation results of CuCr60 with different data size. \textbf{(A)} CuCr60 at \SI{1}{\um} indentation depth with 366 data points. \textbf{(B)} CuCr60 at \SI{1}{\um} depth with 574 data points}\label{fig:cross-validation}
\end{figure}

\cref{fig:cross-validation}$\textbf{A}$ shows the scores for CuCr60 obtained through $k$-fold cross-validation using two clustering algorithms with 366 datapoints at \SI{1}{\um} depth. GMM outperformed KMeans in predicting individual phase content of the CuCr composite. \cref{fig:cross-validation}$\textbf{B}$ shows the scores for varying data sizes after the $k$-fold cross-validation ($k=2,3,...,7$) with the GMM model for CuCr60 data  at \SI{1}{\um} depth. The total number of datasets was increased in increments of 50 until 550 data sets were reached. The standard deviation and mean score of the model decreased as the number of valid datapoints increased, indicating improved accuracy. The model's fit accuracy was considered reliable only when the lowest score was more than 0.95 and the standard deviation smaller than 0.05 in our case. This observation explains the higher accuracy of the CuCr60 results using 574 data compared to CuCr60 with 366 data. 
confirming the viability of the two-dimensional GMM model for predicting the Cr content.

\begin{figure}[ht!]
\centering
\includegraphics[width=0.95\textwidth]{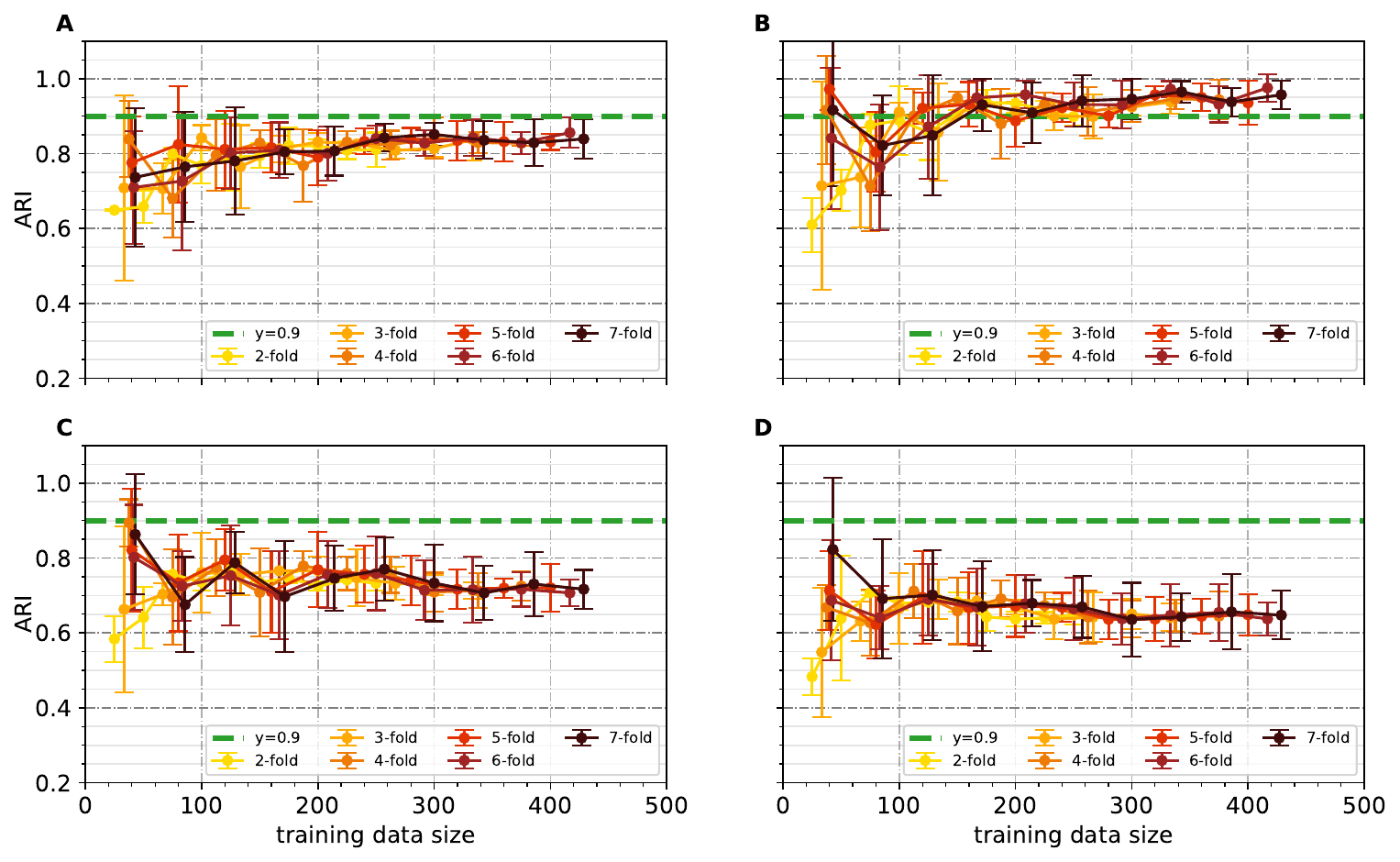}
\caption{Cross-validation results of CuCr25 at \SI{1}{\um} depth with different data size (in total 513 data points). \textbf{(A)} k=2 \textbf{(B)} k=3 \textbf{(C)} k=4 \textbf{(D)} k=5}\label{fig:CuCr25_different_K}
\end{figure}

{\cref{fig:CuCr25_different_K} shows the outcomes of $k$-fold cross-validation of CuCr25 with different clusters. Based on the 2D GMM analysis described above, the optimal model was $k=3$.
However, we also notice that the BIC values determined for $k=4$ and $k=5$ are comparable to those determined for $k=3$ (\cref{fig:CuCr_2D-GMM}$\textbf{C}$). When the BIC values do not indicate a significant difference, the $k$-fold cross-validation can certainly provide a hint as to which model is superior. In the case of $k=3$, the adjusted Rand scores increased with the amount of data, whereas with other numbers of clusters the scores did not exhibit an upward trend in conjunction with predicted scores lower than 80\%. As a supplement to BIC, the $k$-fold cross-validation method can be used to determine the number of clusters and, more importantly, whether the amount of data is sufficient for training.} 

\section{Conclusion}
\label{sec4}

Cu-Cr composites were studied by indentation as a model material to evaluate the ability to determine the properties of individual phases as well as the number of phases present.
1D and 2D Gaussian mixture models were trained and the most likely number of components identified based on the BIC. Using cross-validation we could show that the GMM gave more accurate results than Kmeans clustering. Investigating the dependence of the cross- validation results on the size of the datasets helped to understand what a reasonable amount of data for the training of such models might be.

\appendix
\section{Appendix}
\label{appendix: MLE}
\subsection{Formulation of the Maximum Likelihood Estimate for the Gaussian Model}
In the following, we summarize the most important equations and relations required for 
the BIC criterion for the one-dimensional case. The generalization towards higher 
dimensions follows the same arguments, cf. \citep{neath2012:CS4}. Given a mixture of $k$ 
Gaussian distributions, we start from \cref{eq:PDF} and use it to measure the likelihood 
of a single data point $x_i$:
\begin{equation}\label{eq:likelihood of a single data point}
p\left(x_i \mid \BPhi\right)=\sum_{j=1}^k \alpha_j \mathcal{N}\left(x_i \mid \mu_j, \sigma_j\right)\;,
\end{equation}
where $\BPhi$ is a vector that contains for each component the weight $\alpha_j$ and the Gaussian parameter $\Btheta_j=\{\mu_j, \sigma_j\}$. With this we can define 
the likelihood function \calL\ of the entire data set \calD\ as the 
product of probabilities for each individual value:\color{black}
\begin{equation}\label{eq:Likelihood}
\calL = \prod_{i=1}^N p\left(x_i \mid \BPhi\right)
\end{equation}
and $x_i = \{x_1, ..., x_N\}$ are the data samples. 
$\calL$ acts as the joint probability resulting from the likelihood of all individual data points and is a function of the data and the model. 
Since the data set is fixed, the variables on which $\calL$ depends are therefore the model parameters $\BPhi$. 
The optimal parameters are those that maximize the likelihood of the underlying Gaussian model 
for the given data set
\begin{align}
    \Bmu^\ast = \operatorname{argmax}\limits_\mu \calL(\calD|\Btheta)
    \qquad\text{and}\qquad 
    \boldsymbol{\sigma}^\ast = \operatorname{argmax}\limits_{\boldsymbol{\sigma}} \calL(\calD|\Btheta)\;.
\end{align}
For finding the maxima of \calL, the partial derivative of $\ln{L}$ with respect to each parameter are calculated
\begin{equation}\label{eq:derivation of L}
\frac{\partial \ln \calL}{\partial \mu_j}=0, \frac{\partial \ln \calL}{\partial \sigma_j}=0, \text { and } \frac{\partial \ln \calL}{\partial \alpha_j}=0 \quad \text { where } \quad \ln \mathcal{L}=\sum_{i=1}^N \ln \left[\sum_{j=1}^k \alpha_j \mathcal{N}\left(x_i \mid \mu_j, \sigma_j\right)\right]
\end{equation}

To solve this system of equations, the  expectation maximization (EM) algorithm is used. 
First, we assume that the model parameters ($\mu_j, \sigma_j$) are fixed
for each specific model number $j$ (note that a model $j$ represents a cluster $j$). 
The probability that data $x_i$ belongs to cluster $j$ is denoted by $p\left( j \mid x_i \right)$. 
An often used way to obtain the maximum is to take the derivative of the 
logarithm of the likelihood (note that this does not change the location of 
the maximum as the logarithm function increases monotonically):
\begin{equation}
\frac{\partial \ln \calL}{\partial \theta_j}=-\sum_{i=1}^N p\left( j \mid x_i \right) \frac{\partial}{\partial \theta_j}\left[\ln \sigma_j+\frac{\left(x_i-\mu_j\right)^2}{2 \sigma_j^2}\right]
\end{equation}
By setting each of the derivatives to 0, we obtain the new estimators:

%
\begin{align}\label{eq: new eastimator mu and sigma}
\mu_j=\frac{\sum_{i=1}^N p\left( j \mid x_i \right) x_i}{\sum_{i=1}^N p\left(j \mid x_i\right)}
\qquad&\text{and}\qquad
\sigma_j^2=\frac{\sum_{i=1}^N p\left( j \mid x_i \right)\left(x_i-\mu_j\right)^2}{\sum_{i=1}^N p\left(j \mid x_i\right)}\\
\label{eq: new eastimator alpha}
\text{with}\qquad \alpha_j&=\frac{1}{N} \sum_{i=1}^N p\left( j \mid x_i \right)\;.
\end{align}

In the EM algorithm, the estimation step (E) refers to the calculation and update of  
$p\left( j \mid x_i \right)$ in each iteration, whereas the maximization step 
(M) refers to the calculation (\cref{eq: new eastimator mu and sigma} and 
\cref{eq: new eastimator alpha}) of the updated model parameters, which approach 
the local maximum until convergence. 

\bibliographystyle{Frontiers-Harvard} 
\bibliography{frontiers}

\begin{thebibliography}{24}
\providecommand{\natexlab}[1]{#1}
\expandafter\ifx\csname urlstyle\endcsname\relax
  \providecommand{\doi}[1]{doi:\discretionary{}{}{}#1}\else
  \providecommand{\doi}{doi:\discretionary{}{}{}\begingroup
  \urlstyle{rm}\Url}\fi
\providecommand{\selectlanguage}[1]{\relax}
\providecommand{\bibAnnoteFile}[1]{%
  \IfFileExists{#1}{\begin{quotation}\noindent\textsc{Key:} #1\\
  \textsc{Annotation:}\ \input{#1}\end{quotation}}{}}
\providecommand{\bibAnnote}[2]{%
  \begin{quotation}\noindent\textsc{Key:} #1\\
  \textsc{Annotation:}\ #2\end{quotation}}

\bibitem[{Bishop and Nasrabadi(2006)}]{bishop2006:S4}
Bishop, C.~M. and Nasrabadi, N.~M. (2006).
\newblock \emph{Pattern recognition and machine learning}, vol.~4 (Springer)
\bibAnnoteFile{bishop2006:S4}

\bibitem[{Bos(2019)}]{bos2019:KITthesis}
Bos, C. (2019).
\newblock \emph{Micromechanical characterization of heterogeneous materials,
  statistical analysis of nanoindentation data}.
\newblock Ph.D. thesis, Karlsruhe Institute of Technology (KIT)
\bibAnnoteFile{bos2019:KITthesis}

\bibitem[{Chac{\'o}n and Rastrojo(2023)}]{chacon2023:ADAC17}
Chac{\'o}n, J.~E. and Rastrojo, A.~I. (2023).
\newblock Minimum adjusted rand index for two clusterings of a given size.
\newblock \emph{Advances in Data Analysis and Classification} 17, 125--133
\bibAnnoteFile{chacon2023:ADAC17}

\bibitem[{Chakrabarti and Laughlin(1984)}]{Chakrabarti1984:BullAllPhaDiag}
Chakrabarti, D.~J. and Laughlin, D.~E. (1984).
\newblock The cr-cu (chromium-copper) system.
\newblock \emph{Bulletin of Alloy Phase Diagrams} 5, 59--68
\bibAnnoteFile{Chakrabarti1984:BullAllPhaDiag}

\bibitem[{De~Backer et~al.(2013)De~Backer, Martinez, Rosenauer, and
  Van~Aert}]{de2013:U134}
De~Backer, A., Martinez, G.~T., Rosenauer, A., and Van~Aert, S. (2013).
\newblock Atom counting in haadf stem using a statistical model-based approach:
  Methodology, possibilities, and inherent limitations.
\newblock \emph{Ultramicroscopy} 134, 23--33
\bibAnnoteFile{de2013:U134}

\bibitem[{Gideon(1978)}]{gideon1978:TAS6}
Gideon, S. (1978).
\newblock Estimating the dimension of a model.
\newblock \emph{The Annals of Statistics} 6, 461
\bibAnnoteFile{gideon1978:TAS6}

\bibitem[{Jacob et~al.(2000)Jacob, Priya, and Waseda}]{Jacob2000:ZMetall}
Jacob, K., Priya, S., and Waseda, Y. (2000).
\newblock Thermodynamic study of liquid cu-cr alloys and metastable liquid
  immiscibility.
\newblock \emph{Zeitschrift für Metallkunde/Materials Research and Advanced
  Techniques} 91, 594–600
\bibAnnoteFile{Jacob2000:ZMetall}

\bibitem[{Kossman and Bigerelle(2021)}]{kossman2021:M14}
Kossman, S. and Bigerelle, M. (2021).
\newblock Pop-in identification in nanoindentation curves with deep learning
  algorithms.
\newblock \emph{Materials} 14, 7027
\bibAnnoteFile{kossman2021:M14}

\bibitem[{Lebedev et~al.(1995)Lebedev, Berunkov, Romanov, Kopylov, Filonenko,
  and Gryaznov}]{Lebedev1995:MSEA203}
Lebedev, A., Berunkov, Y., Romanov, A., Kopylov, V., Filonenko, V., and
  Gryaznov, V. (1995).
\newblock Softening of the elastic modulus in submicrocrystalline copper.
\newblock \emph{Materials Science and Engineering A} 203, 165--170
\bibAnnoteFile{Lebedev1995:MSEA203}

\bibitem[{Liu et~al.(2018)Liu, Ostadhassan, and Bubach}]{liu2018:JPSE167}
Liu, K., Ostadhassan, M., and Bubach, B. (2018).
\newblock Application of nanoindentation to characterize creep behavior of oil
  shales.
\newblock \emph{Journal of Petroleum Science and Engineering} 167, 729--736
\bibAnnoteFile{liu2018:JPSE167}

\bibitem[{Neath and Cavanaugh(2012)}]{neath2012:CS4}
Neath, A.~A. and Cavanaugh, J.~E. (2012).
\newblock The bayesian information criterion: background, derivation, and
  applications.
\newblock \emph{Wiley Interdisciplinary Reviews: Computational Statistics} 4,
  199--203
\bibAnnoteFile{neath2012:CS4}

\bibitem[{Oliver and Pharr(1992)}]{oliver1992:JMR7}
Oliver, W.~C. and Pharr, G.~M. (1992).
\newblock An improved technique for determining hardness and elastic modulus
  using load and displacement sensing indentation experiments.
\newblock \emph{Journal of Materials Research} 7, 1564--1583
\bibAnnoteFile{oliver1992:JMR7}

\bibitem[{Oliver and Pharr(2004)}]{oliver2004:JMR19}
Oliver, W.~C. and Pharr, G.~M. (2004).
\newblock Measurement of hardness and elastic modulus by instrumented
  indentation: Advances in understanding and refinements to methodology.
\newblock \emph{Journal of Materials Research} 19, 3--20
\bibAnnoteFile{oliver2004:JMR19}

\bibitem[{{\"O}ztuna et~al.(2006){\"O}ztuna, Elhan, and
  T{\"u}ccar}]{oztuna2006:TJMS36}
{\"O}ztuna, D., Elhan, A.~H., and T{\"u}ccar, E. (2006).
\newblock Investigation of four different normality tests in terms of type 1
  error rate and power under different distributions.
\newblock \emph{Turkish Journal of Medical Sciences} 36, 171--176
\bibAnnoteFile{oztuna2006:TJMS36}

\bibitem[{Pedregosa et~al.(2011)Pedregosa, Varoquaux, Gramfort, Michel,
  Thirion, Grisel et~al.}]{scikit-learn2011:JMLR}
Pedregosa, F., Varoquaux, G., Gramfort, A., Michel, V., Thirion, B., Grisel,
  O., et~al. (2011).
\newblock Scikit-learn: Machine learning in python.
\newblock \emph{the Journal of machine Learning research} 12, 2825--2830
\bibAnnoteFile{scikit-learn2011:JMLR}

\bibitem[{Prakash and Sandfeld(2022)}]{prakash2022:AEM24}
Prakash, A. and Sandfeld, S. (2022).
\newblock Automated analysis of continuum fields from atomistic simulations
  using statistical machine learning.
\newblock \emph{Advanced Engineering Materials} 24, 2200574
\bibAnnoteFile{prakash2022:AEM24}

\bibitem[{Randall et~al.(2009)Randall, Vandamme, and Ulm}]{randall2009:JMR24}
Randall, N.~X., Vandamme, M., and Ulm, F.-J. (2009).
\newblock Nanoindentation analysis as a two-dimensional tool for mapping the
  mechanical properties of complex surfaces.
\newblock \emph{Journal of Materials Research} 24, 679--690
\bibAnnoteFile{randall2009:JMR24}

\bibitem[{Reynolds and Rose(1995)}]{reynolds1995:ITSAP3}
Reynolds, D.~A. and Rose, R.~C. (1995).
\newblock Robust text-independent speaker identification using gaussian mixture
  speaker models.
\newblock \emph{IEEE Transactions on Speech and Audio Processing} 3, 72--83
\bibAnnoteFile{reynolds1995:ITSAP3}

\bibitem[{Reynolds et~al.(2009)}]{reynolds2009:EB741}
Reynolds, D.~A. et~al. (2009).
\newblock Gaussian mixture models.
\newblock \emph{Encyclopedia of Biometrics} 741
\bibAnnoteFile{reynolds2009:EB741}

\bibitem[{Sorelli et~al.(2008)Sorelli, Constantinides, Ulm, and
  Toutlemonde}]{sorelli2008:CCR38}
Sorelli, L., Constantinides, G., Ulm, F.-J., and Toutlemonde, F. (2008).
\newblock The nano-mechanical signature of ultra high performance concrete by
  statistical nanoindentation techniques.
\newblock \emph{Cement and Concrete Research} 38, 1447--1456
\bibAnnoteFile{sorelli2008:CCR38}

\bibitem[{Venderley et~al.(2022)Venderley, Mallayya, Matty, Krogstad, Ruff,
  Pleiss et~al.}]{venderley2022:PNAS119}
Venderley, J., Mallayya, K., Matty, M., Krogstad, M., Ruff, J., Pleiss, G.,
  et~al. (2022).
\newblock Harnessing interpretable and unsupervised machine learning to address
  big data from modern x-ray diffraction.
\newblock \emph{Proceedings of the National Academy of Sciences} 119,
  e2109665119
\bibAnnoteFile{venderley2022:PNAS119}

\bibitem[{Vignesh et~al.(2019)Vignesh, Oliver, Kumar, and
  Phani}]{vignesh2019:MD181}
Vignesh, B., Oliver, W., Kumar, G.~S., and Phani, P.~S. (2019).
\newblock Critical assessment of high speed nanoindentation mapping technique
  and data deconvolution on thermal barrier coatings.
\newblock \emph{Materials \& Design} 181, 108084
\bibAnnoteFile{vignesh2019:MD181}

\bibitem[{von Klinski-Berger(2015)}]{vonKlinski2015:TUDthesis}
von Klinski-Berger, K. (2015).
\newblock \emph{Charakterisierung von Kupfer-Chrom-Verbundwerkstoffen f{\"u}r
  die Schalttechnik}.
\newblock Ph.D. thesis, Technische Universit{\"a}t Darmstadt
\bibAnnoteFile{vonKlinski2015:TUDthesis}

\bibitem[{Yeo et~al.(2023)Yeo, Shigematsu, and Katori}]{yeo2023:JSG}
Yeo, T., Shigematsu, N., and Katori, T. (2023).
\newblock Dynamically recrystallized grains identified via the application of
  gaussian mixture model to ebsd data.
\newblock \emph{Journal of Structural Geology} , 104800
\bibAnnoteFile{yeo2023:JSG}

\end{thebibliography}

\end{document}